\documentclass[11pt]{article}

\usepackage{acl2014}
\usepackage{times,amsmath,amsthm}
\usepackage{latexsym}
\usepackage{amsmath}
\usepackage{multirow}
\usepackage{url}
\usepackage{mathtools}
\usepackage{hhline}
\usepackage{qtree}
\usepackage{amsmath}
\usepackage{amsfonts}
\usepackage{xcolor}

\author{
Chiraag Lala \\
Out of The View Ventures \\
London, UK \\ 
{\tt chiraag.r.lala@gmail.com} \\ \And
Shay B. Cohen \\
School of Informatics \\
University of Edinburgh, Edinburgh, UK \\
{\tt scohen@inf.ed.ac.uk}
}

\title{The Visualization of Change in Word Meaning over Time using Temporal Word Embeddings}

\begin{document}

\maketitle

\begin{abstract}
We describe a visualization tool that can be used to view the change in meaning of words over time. The tool makes use of existing (static) word embedding datasets
together with a timestamped $n$-gram corpus to create {\em temporal} word embeddings.
\end{abstract}

\section{Introduction}

Embedding words into a vector space and using these vectors in various applications has been a recent topic of great interest in NLP
\cite{turian2010word,dhillon2011multi,collobert2011natural,mikolov2011rnnlm,blacoe2012comparison,faruqui2014improving}.
Many of these methods rely on dimensionality reduction techniques for vectors which represent distributional features of the words.

In this short paper, we describe an approach to visualize the change in meaning of words over time using such word embeddings. The main idea is to embed words, given as a query,
into a vector space, in a running sequence of time slices. Once these embeddings are calculated, they can be represented on a 2D plane.
The points in the 2D plane move in the space as the time slices change.

Assuming that word vectors close to each other in Euclidean space are semantically related, and assuming that distributional similarity and context of appearance
of words also greatly determines the meaning of the word \cite{firth-61}, our visualization tool can be viewed as a tool to inspect the change in word meaning over time.

\section{Temporal Word Embeddings}

Let $V$ be a vocabulary -- a set of words over some alphabet. We also assume a special symbol $\bot \in V$, which denotes an unknown word token.
For example, in our online visualization tool, $V$ is a subset of the words that appear in the
Google books n-gram corpus \cite{michel2011quantitative}. We define a static word embedding function to be function $f \colon V \rightarrow \mathbb{R}^d$, which maps every word in $V$
to a vector in some $d$-dimensional space. We experimented with several such word embedding functions, including those by \newcite{collobert2011natural} that
make use of neural networks with the SENNA toolkit, the HLBL embeddings of \newcite{turian2010word} and \newcite{mnih2007three} and also the embeddings by \newcite{mikolov2011rnnlm} that make use of the RNNLM toolkit.

Let $T$ be a set of time slices. Each member $t \in T$ denotes a certain span of time, for example, the years between 1810 and 1815.
A temporal word embedding function for $T$ and $V$ is then a function $g \colon T \times V \rightarrow \mathbb{R}^{\ell}$. In our visualization tool, $T$ denotes the span
of years between 1800 and 2008, broken into 5 years intervals. We assume a function $\mathrm{TimeSlice}$ that maps an arbitrary timestamp $t$ into its corresponding
time slice in $T$. For example, in our experiments $\mathrm{TimeSlice}(1801)$ would map the timestamp 1801 to the range of years 1800-1805.

We now show how to convert a static word embedding function to a temporal word embedding function. We first assume an associative commutative operator $\odot$ which operates on a pair
of vectors. This operator takes as input a pair of vectors (not necessarily of the same dimension) and returns a new vector. In our visualization tool, $\odot$ is the
concatenation operator between vectors, and also the sum of vectors. In the latter case, we assume that the operator $\odot$ always accepts vectors of the same length.

In order to create the temporal word embedding, we assume the existence of a dataset $D$. Each datum $d \in D$ is an $n$-gram ($n = 5$ in our case and in general
should be odd, so a middle word can be identified), together with a timestamp $t$ in which this $n$-gram appeared and a count for it $c$ in a corpus from that time. Therefore, $d = \langle w_1,\ldots,w_n, t, c \rangle$.

In order to convert $f$ to a temporal word embedding $g$, we define $g(t,w)$ to be:

\begin{align}
g(t,w) = \hspace{-0.19in} \sum_{\substack{d = \langle w_1,\ldots,w_n, t', c \rangle \in D \\ \mathrm{TimeSlice}(t') = t, w_{(n+1)/2} = w }} \hspace{-0.5in} (c/N_{t,w}) \times (\bigodot_{i \neq (n+1)/2} \hspace{-0.1in} f(w_i)),
\end{align}

\noindent where $N_{t,w}$ is defined as:

\begin{align}
N_{t,w} = \sum_{\substack{d = \langle w_1,\ldots,w_n, t', c \rangle \in D \\ \mathrm{TimeSlice}(t') = t, w_{(n+1)/2} = w }} c
\end{align}

This means that $g(t,w)$ is calculated by running the operator $\odot$ over all contexts of the word $w$ in $D$ (meaning, words to left and words to right) in time slice $t$.

\paragraph{Data} For the development of the temporal word embeddings, we used a subset of the Google books 5-gram data \cite{michel2011quantitative} for years 1800-2008.

\section{Multidimensional Scaling}

Once we obtain the temporal function $g(t,w)$, we use it to embed a collection of words in a query into the plane, and have a smooth animation that moves the words
through the time slices.

For a given query $w_1,\ldots,w_k$, we compute $g(t,w_i)$ for all $i \in \{1 ,\ldots,k\}$ and $t \in T$. Let $v_{t,i} = g(t,w_i)$.
We then create a matrix $A$ of size $k|T| \times k|T|$, where $$A_{(i,t),(j,t')} = || v_{t,i} - v_{t',j}||_2.$$ The matrix $A$ is therefore the
distance matrix for all vectors across all time slices for all words in the query.

We then use multidimensional scaling with the distance matrix $A$ to embed all points $v_{t,i}$ in a 2D plane. Multidimensional scaling works
by solving the following optimization problem over the variables $x_{t,i} \in \mathbb{R}^2$ for $t \in T$ and $i \in \{ 1,\ldots,k \}$:

\begin{align}
\min_{x_{t,i} \colon t \in T, i \in \{1,\ldots,k\}} \sum_{t,i,t',j} \left( ||x_{t,i} - x_{t',j} ||_2 - A_{(t,i),(t',j)} \right)^2.
\end{align}

It can be solved using eigenvalue decomposition as a blackbox, that is run on a matrix which is a transformed version of $A$. See \newcite{borg2005modern} for more details.

\section{Putting It All Together}

Given a query of several words, $w_1,\ldots,w_k$, we compute a two-dimensional word embedding for each word in each time slice.
We then plot these embeddings through time, moving each word through the time slices using a basic line drawing algorithm based on
\newcite{bresenham1965algorithm}.

The final tool we developed can be viewed online at \url{http://kinloch.inf.ed.ac.uk/words}.

\section{Future Work}

We hope to make this visualization tool useful for researchers who do diachronic analysis of text or words.
In addition, we believe that the idea of temporal word embeddings is useful for various classification tasks,
in which there is a temporal component to the data. For example, such temporal word embeddings can
assist in classifying documents into the year they were written in. Preliminary work done by the first author
as part of his Master's project shows that indeed this is the case, and temporal word embeddings can be used
as features for a diachronic classification problem.

\section*{Acknowledgements}

We would like to thank Mirella Lapata and Bonnie Webber for useful feedback on this tool.
This tool was created as part of the first author's M.Sc. project at the University of Edinburgh, School
of Informatics.

\bibliographystyle{acl}
\bibliography{nlp}

\begin{thebibliography}{}

\bibitem[\protect\citename{Blacoe and Lapata}2012]{blacoe2012comparison}
W.~Blacoe and M.~Lapata.
\newblock 2012.
\newblock A comparison of vector-based representations for semantic
  composition.
\newblock In {\em Proceedings of the 2012 Joint Conference on Empirical Methods
  in Natural Language Processing and Computational Natural Language Learning},
  pages 546--556. Association for Computational Linguistics.

\bibitem[\protect\citename{Borg and Groenen}2005]{borg2005modern}
I.~Borg and P.~J.~F. Groenen.
\newblock 2005.
\newblock {\em Modern multidimensional scaling: Theory and applications}.
\newblock Springer.

\bibitem[\protect\citename{Bresenham}1965]{bresenham1965algorithm}
J.~E. Bresenham.
\newblock 1965.
\newblock Algorithm for computer control of a digital plotter.
\newblock {\em IBM Systems journal}, 4(1):25--30.

\bibitem[\protect\citename{Collobert \bgroup et al.\egroup
  }2011]{collobert2011natural}
R.~Collobert, J.~Weston, L.~Bottou, M.~Karlen, K.~Kavukcuoglu, and P.~Kuksa.
\newblock 2011.
\newblock Natural language processing (almost) from scratch.
\newblock {\em The Journal of Machine Learning Research}, 12:2493--2537.

\bibitem[\protect\citename{Dhillon \bgroup et al.\egroup
  }2011]{dhillon2011multi}
P.~Dhillon, D.~P. Foster, and L.~H. Ungar.
\newblock 2011.
\newblock Multi-view learning of word embeddings via cca.
\newblock In {\em Advances in Neural Information Processing Systems}, pages
  199--207.

\bibitem[\protect\citename{Faruqui and Dyer}2014]{faruqui2014improving}
M.~Faruqui and C.~Dyer.
\newblock 2014.
\newblock Improving vector space word representations using multilingual
  correlation.
\newblock {\em Proc. of EACL. Association for Computational Linguistics}.

\bibitem[\protect\citename{Firth}1961]{firth-61}
John~Rupert Firth.
\newblock 1961.
\newblock {\em Papers in Linguistics 1934-1951: Repr}.
\newblock Oxford University Press.

\bibitem[\protect\citename{Michel \bgroup et al.\egroup
  }2011]{michel2011quantitative}
J.-. Michel, Y.~K. Shen, A.~P. Aiden, A.~Veres, M.~K. Gray, J.~P. Pickett,
  D.~Hoiberg, D.~Clancy, P.~Norvig, J.~Orwant, et~al.
\newblock 2011.
\newblock Quantitative analysis of culture using millions of digitized books.
\newblock {\em science}, 331(6014):176--182.

\bibitem[\protect\citename{Mikolov \bgroup et al.\egroup
  }2011]{mikolov2011rnnlm}
Tomas Mikolov, Stefan Kombrink, Anoop Deoras, Lukar Burget, and J~Cernocky.
\newblock 2011.
\newblock Rnnlm-recurrent neural network language modeling toolkit.
\newblock In {\em Proc. of the 2011 ASRU Workshop}, pages 196--201.

\bibitem[\protect\citename{Mnih and Hinton}2007]{mnih2007three}
A.~Mnih and G.~Hinton.
\newblock 2007.
\newblock Three new graphical models for statistical language modelling.
\newblock In {\em Proceedings of the 24th international conference on Machine
  learning}, pages 641--648. ACM.

\bibitem[\protect\citename{Turian \bgroup et al.\egroup }2010]{turian2010word}
J.~Turian, L.~Ratinov, and Y.~Bengio.
\newblock 2010.
\newblock Word representations: a simple and general method for semi-supervised
  learning.
\newblock In {\em Proceedings of the 48th Annual Meeting of the Association for
  Computational Linguistics}, pages 384--394. Association for Computational
  Linguistics.

\end{thebibliography}

\end{document}